\documentclass[runningheads]{llncs}
\usepackage[T1]{fontenc}
\usepackage{graphicx}
\usepackage{booktabs}
\usepackage[misc]{ifsym}
\newcommand{\corr}{(\Letter)}

\usepackage{xurl}
\usepackage{hyperref}
\usepackage{amsmath}
\usepackage{amssymb}
\usepackage{bbm}
\usepackage{multirow} 
\usepackage{siunitx}
\usepackage{rotating}

\begin{document}

\title{Grokipedia \textit{vs} Wikipedia: An LLM-Based Audit of Political Neutrality along Ideologies}

\titlerunning{Grokipedia \textit{vs} Wikipedia: An LLM-Based Audit of Political Neutrality}

\author{Filippos Vlahos \inst{1}\corr \and
Guillaume Bied\inst{1} \and
Tijl De Bie \inst{1}}

\authorrunning{F. Vlahos et al.}

\institute{Ghent University, BE, \email{\{filippos.vlahos\}@ugent.be}}

\maketitle              

\begin{abstract}
Online encyclopedias shape political opinion and, through it, democratic discourse. In late 2025, Grokipedia was released, an encyclopedia written entirely by the LLM Grok. One motivation behind the project was to provide an unbiased alternative to Wikipedia, which has faced accusations of ``left-wing'' and ``liberal'' bias. But does an encyclopedia written by an LLM deliver greater neutrality, or does it simply embed a different ideology? We conduct a large-scale political bias study on Grokipedia and Wikipedia, analysing 1,394 article pairs describing members of government for neutrality along nine expert-coded ideology dimensions employing four LLM judges, Grok, Claude, Mistral, and DeepSeek. As the LLMs could themselves be biased, we also investigate patterns in their judgments. We find all LLM-judges, including Grok, to rate Grokipedia less neutral than Wikipedia. Both encyclopedias are rated as portraying politicians favourably overall, but towards different ideological groups. Grokipedia particularly favours economically right-wing politicians and penalises socially liberal ones, while Wikipedia is rated as favourably biased towards the latter.

\end{abstract}

\section{Introduction}

\begin{figure*}[t]
    \centering
    \includegraphics[width=\linewidth]{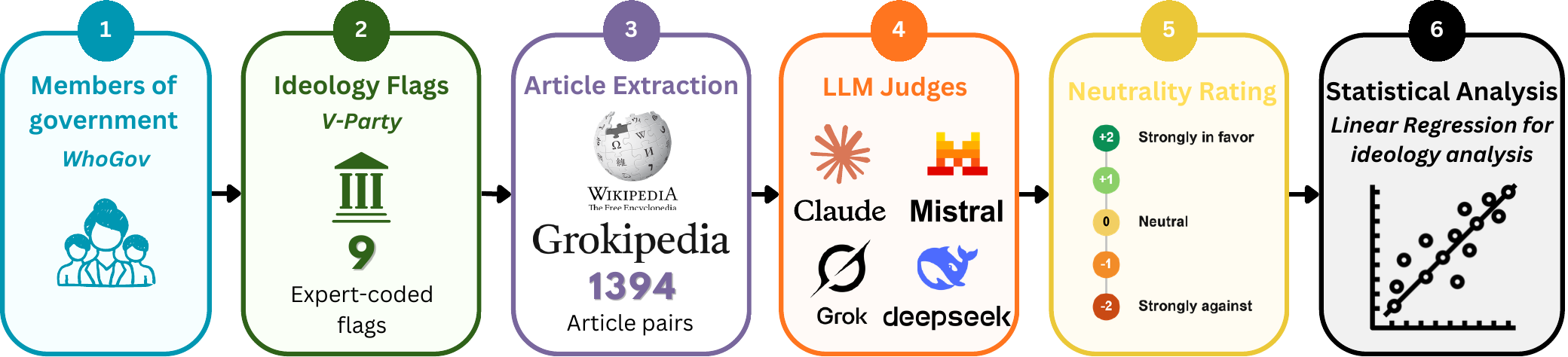}
    \caption{Methodology overview: members of government (\textit{WhoGov}, 2016–2023) are mapped to expert-coded V-Party ideology scores, their Wikipedia and Grokipedia articles are rated for neutrality by four LLM judges, and the effects of ideology on the ratings are estimated via OLS regression.}
    \label{fig:methodology}
\end{figure*}

Online encyclopedias are important components of today's democracies as they serve as public knowledge bases for a broad range of topics including history, science, and politics. Traditionally, they are written by humans, with Wikipedia being by far the most prominent example, receiving on the order of hundreds of millions of page views per day.\footnote{Based on Wikimedia Foundation traffic statistics; see also \cite{pew_wikipedia_traffic_2026}.}

In October 2025, \textit{xAI} launched Grokipedia, a new kind of free online encyclopedia entirely generated by the Grok Large Language Model (LLM) \cite{grok4_model}. As of November 2025, Grokipedia is on version 0.2, and includes over a million articles. 
Grokipedia was positioned by xAI as a step towards its goal of better ``understanding the Universe'' as well as a more neutral alternative to Wikipedia, which xAI's founder Elon Musk has characterised as ``Wokepedia'' \cite{musk2024wokepedia} and ``an extension of legacy media propaganda'' \cite{musk2025propaganda}.

Grokipedia has not yet achieved traffic comparable to Wikipedia~\cite{pcmag2025grokipedia}, although it is still in beta with a first major release ahead. Beyond direct traffic, it could reach a wide audience indirectly as reports have shown LLMs citing it~\cite{down2026latest} with future LLMs potentially being trained on its contents. More generally, the examples of Grokipedia and other recent LLM encyclopedias~\cite{saeed2026llmpedia} indicate that with enough compute and resources, governments, companies or other actors can rapidly create and distribute extensive knowledge bases, with the risk of such initiatives being used to embed and promote particular ideologies. Therefore, devising methods to rapidly audit the ideological positioning of such encyclopedias at scale is of direct relevance to the integrity of public information ecosystems.

This work addresses the research question: \textbf{How do Grokipedia and Wiki-\\pedia compare in terms of political neutrality, and which ideologies does each platform systematically favour or penalise?}

We address this question by introducing a methodology for analysing political bias at scale in articles on members of government, visualised in Fig.~\ref{fig:methodology}. We first assemble a list of members of government, and map them along nine ideology dimensions using the expert-coded V-Party dataset. We then employ a panel of four LLM-judges, Claude, Grok, Mistral, and DeepSeek to evaluate each article on neutrality. Finally, we analyse the results using two linear regression specifications to estimate the effects of the ideologies on perceived neutrality.

The methodology gives rise to a secondary question which we investigate, concerning the role of the evaluator itself: \textbf{To what extent does the choice of LLM judge influence neutrality assessments?}

We make no claims that our measure of bias is absolute, and acknowledge the limitations of our operationalisation of neutrality assessments. To mitigate the risks inherent in the use of LLM judges, we employ LLMs with different documented ideological tendencies, allowing us to study their biases and partially separate them from those exhibited in the encyclopedias.

We find that both encyclopedias contain a substantial share of articles rated as biased by the LLMs. Of the two, Grokipedia contains more biased-rated articles in total. Towards most politicians, and especially those on the economic right, this bias is favourable, while it is less favourable towards those holding socially liberal views. Conversely, Wikipedia is rated more biased in favour of socially liberal politicians, but is deemed more neutral overall. All four LLMs are aligned more closely to Wikipedia than to Grokipedia, a remarkable result given that Grok is the LLM behind Grokipedia. 

The remainder of this paper is structured as follows. Section~\ref{sec:related_work} reviews related work on political bias in Wikipedia and prior studies on Grokipedia. Section~\ref{sec:methodology} describes our dataset construction, LLM judge selection, and OLS specifications. Section~\ref{sec:results} presents descriptive statistics and regression results. Section~\ref{sec:discussion} interprets these findings, discusses the evaluative tendencies of the LLM judges, and addresses the implications of Grok rating its own content as biased. Section~\ref{sec:conclusion} concludes with limitations and directions for future work.

\section{Related Work}\label{sec:related_work}

\paragraph{Political Bias on Wikipedia} Studies of political bias in Wikipedia have shown that although the platform overall displays neutral tendencies, those are less prominent in political articles. Many articles on US politics have been found to contain bias with both the level and direction of that bias evolving over time~\cite{greenstein2012wikipedia}. More specifically, a mild to moderate tendency in Wikipedia articles to associate right-of-centre public figures with more negative sentiment than left-of-centre ones has been observed~\cite{rozado2024wikipedia}. A common source of bias in Wikipedia has been argued~\cite{ackerly2022wikipedia} to stem from the limited diversity of its editors, which predominantly belong to particular groups (e.g. Western, young, white, and male) as well as of the cited sources which are often authored by similar groups. 

\paragraph{Wikipedia \textit{vs} Grokipedia}
Research on Grokipedia has so far focused primarily on comparisons to Wikipedia. Using quantitative analyses, Grokipedia has been found to be a ``synthetic derivative''~\cite{yasseri2025similar} of Wikipedia, but with significantly longer articles~\cite{triedman2025did,yasseri2025similar}. In addition, studies report citation quality is poorer in Grokipedia than in Wikipedia \cite{mehdizadeh2025epistemic,triedman2025did}. These issues are amplified for political and controversial articles, where unreliable source use increases further and Grokipedia diverges more from Wikipedia than average \cite{hadad2026wikipedia,triedman2025did}. That this divergence is concentrated in political and controversial content may indicate these are the areas where particular editorial attention has been directed in Grokipedia's development.

Two studies have focused on comparing political ideology between Grokipedia and Wikipedia. The first~\cite{eibl2026grokipedia} used  a \textit{RoBERTa}-based classifier to score six article pairs on a left-right axis. They found that both sources predominantly adopt left-leaning framings, but that Grokipedia exhibits a modest but consistent right-leaning bias relative to Wikipedia. The second~\cite{hadad2026wikipedia} used \textit{gemma3:4b} to detect content framing in the introduction of 395 Grokipedia-Wikipedia article pairs about United States (U.S.) politics, geopolitics, and conspiracies. They found that framing patterns in Grokipedia are correlated with those observed in Wikipedia, with the exception of U.S. politics articles, for which systematically more laudatory framing and a reduced emphasis on conflict or controversy relative to Wikipedia were observed.

All previous studies on Grokipedia have been conducted on V0.1 which we found to be much more verbose than V0.2. In the 1,394 articles we examined the average article length drops from around 15,500 words in V0.1 to 6,500 in V0.2. Thus, some past findings may not hold to the same extent in the current version. To the best of our knowledge, ours is the first study examining V0.2. 

Additionally, this is the largest political ideology study comparing Grokipedia and Wikipedia, examining 1,394 article pairs compared to 6~ \cite{eibl2026grokipedia} and 395~\cite{hadad2026wikipedia}. Unlike previous studies we also analyse the entirety of each article and make use of state-of-the-art LLMs. 



\section{Methodology}\label{sec:methodology}


This section is split into two parts, respectively describing (1) the collection of the dataset, and (2) the analysis strategy. Our methodology makes three key contributions: i) it operates at \textbf{scale} across 1,394 politicians, ii) it \textbf{grounds the analysis} in expert-coded ideological dimensions, and, iii) it employs \textbf{four diverse LLM judges} to rate neutrality while accounting for their individual biases.

\subsection{Dataset Construction}

\paragraph{Subject Collection} First, we constructed a corpus of articles likely to exhibit political ideological variation. For this, we used the \textit{WhoGov} dataset \cite{nyrup2020governs} which contains yearly data on members of government in 177 countries. From this dataset we extracted the names of 9,752 members of government who served between 2016 and 2023; these individuals are the subjects of our analysis.

\paragraph{Ideology Dimensions} As our aim is to study the ideological biases within each encyclopedia, we scored politicians from our list along political ideology dimensions to ground our analysis. For this, we used the \textit{V-Party} dataset~\cite{lindberg2022codebook}, which contains expert-coded assessments of party positions on election years for the period 1970–2019.

Using expert coded dimensions distinguishes our approach from related work which use LLMs to generate or annotate ideological scores for political figures and texts \cite{buyl2026large,eibl2026grokipedia}. Relying on LLM-generated ideology scores would be circular; the models whose biases we aim to disentangle would partly define the categories against which those biases are measured. Our approach severs this dependency, adds ecological validity, and ensures that the ideological grounding of our analysis is orthogonal to the LLM judgments we study.

From V-Party, we considered the following nine dimensions, where higher values indicate stronger alignment with the stated position and lower values the opposite:\footnote{Some dimensions from the V-Party dataset were renamed to improve intuitiveness. Original names can be found in Appendix \ref{app:v-party}. More information on the definition of the ideological dimensions can be found in the V-Party codebook.}

\begin{enumerate}
    \item \textbf{Cultural inclusivity} (\textit{culincl}): Opposes cultural superiority of a specific social group or nation. 
    \item \textbf{Immigration openness} (\textit{immig}): Supports immigration into the country.
    \item \textbf{LGBT equality} (\textit{lgbt}): Supports social equality for the LGBT community.
    \item \textbf{Secular positioning} (\textit{secular}): Refrains from invoking religion to justify political positions.
    \item \textbf{Women's labour equality} (\textit{womlab}): Supports equal participation of women in the labour market.
    \item \textbf{Anti-clientelism} (\textit{anti-client}): Refrains from targeted provision of goods and benefits to gain votes.
    \item \textbf{Political pluralism} (\textit{plural}): Shows commitment to democratic norms and free elections.
    \item \textbf{Anti-populism} (\textit{anti-popul}): Refrains from using populist rhetoric.
    \item \textbf{Right-wing} (\textit{rightwing}): Ideological stance on economic issues. Right-wing indicates promoting a reduced economic role for government (e.g. less regulation, and a leaner welfare state).
\end{enumerate}

We mapped each politician to their party's ideology in the year closest to their most recent election, yielding 6,846 politicians paired to the ideology dimensions. This attribution of party-level scores to individuals rests on two premises. First, members of government occupy prominent ideological roles within their party, increasingly serving as the architects and public face of its platform~\cite{garzia2022partisan,poguntke2007presidentialization,wagner2012parties}. Second, core party ideologies change little across electoral cycles, as shifting them carries reputational risks~\cite{adams2004understanding,meijers2025accepts,somer2009timely,werner2025parties}. The mapping is nonetheless imperfect, as individuals may diverge from their party's position or shift ideologically between the annotated election year and the time of study. However, such errors should not deviate systematically towards any particular ideology and, given our sample size, should not materially affect the results.

\paragraph{Article Collection} Subsequently, we collected 1,836 article pairs for politicians from Wikipedia and Grokipedia. The sharp decline in coverage (from 6,846 to 1,836) was largely due to politicians lacking entries in both sources.  We verified this by randomly sampling 100 politicians with no matched article and manually searching for their entries. We confirmed that only 2\% of our sample had qualifying articles we had missed.  Next, we applied an article length lower-bound of 300 words to ensure enough content is present for an informed judgment. This led to another 180 article pairs being dropped. We then dropped another 29 article pairs as they lacked ideology annotations in one or more of the dimensions considered.

As a final check, we verified that retrieved articles corresponded to the intended politicians by searching in the article for the subject's year of birth and for politics-related keywords. This led to further reducing our sample by 233 article pairs that failed these filters. To validate the coherence of the final dataset, we drew a random sample of 100 retained article pairs and manually confirmed that only 1\% did not correspond to the intended politician. This procedure yielded a final dataset of 1,394 politicians, each paired with two articles and nine ideological scores.


\paragraph{LLM Judges} Upon assembling our dataset, we prompted LLMs to assess each article on neutrality. We assembled a panel of four reasoning instruction-tuned LLMs. First, Grok-4~\cite{grok4_model} was selected as a rater given its role as the generative model underlying Grokipedia, enabling us to examine whether it exhibits systematic leniency toward self-generated content. Another LLM selected was Claude-Opus-4.6~\cite{claude_opus_46}, a state-of-the-art model at the time of writing, whose model-family has been shown to achieve the strongest bias-rating performance among commercial LLMs~\cite{kennedy2026left}. In addition, Anthropic, the company behind Claude, have trained it explicitly to be ``politically even-handed''~\cite{anthropic2025evenhandedness}. The third judge selected was DeepSeek-V4-Pro~\cite{deepseek_v4_pro}, which adds ideological diversity as a non-western model~\cite{buyl2026large}. Finally, we included a European model
Mistral-Medium-3.5~\cite{mistral_medium_35}. All models were the state-of-the-art offerings from their respective providers at the time of polling.

The LLM-judges were selected in part to introduce ideological diversity into the evaluation pipeline. Empirical studies have consistently documented a left-of-centre tendency across the majority of frontier LLMs, a pattern in which Claude is representative~\cite{choudhary2024political,peng2026beyond,rozado2024political}. The evidence on Grok is mixed; some evaluations place it within this left-leaning majority~\cite{rozado2024political}, while others identify it as the only right-leaning model among eight prominent LLMs ~\cite{khetan2026politicsbench} and the most ideologically expressive in a nine-model comparison~\cite{kennedy2026left}. DeepSeek introduces a third, orthogonal axis of variation. While it falls within the economically-left, socially-liberal majority~\cite{dormuth2025cautionary}, as a Chinese model its distinctiveness lies on the geopolitical dimension~\cite{buyl2026large}. Mistral introduces a fourth, European axis of variation. Rather than exhibiting a uniform ideological slant, it displays a centrist default alongside highly localised, non-uniform political entity biases~\cite{chen2026uncovering}. Pairing these models thus reduces the risk that outcomes reflect the systematic political homogeneity of most frontier LLMs.

\paragraph{Prompt}
Our task asks each LLM to assess a single article on neutrality. A challenge in assessing neutrality is that the concept admits many interpretations, and leaving it implicit would make the LLM judgments harder to interpret and compare. We therefore formally define neutrality within the prompt\footnote{We provide the exact prompt in Appendix~\ref{apx:Prompt}.}. Our definition is inspired by Wikipedia's \textit{Neutral Point Of View} (NPOV) principle, as it is a widely understood formulation of neutrality for encyclopedias and Grokipedia does not provide an alternative. Concretely, a neutral article should i) present all significant viewpoints fairly and proportionally; ii) avoid taking sides; iii) use fact-focused, non-persuasive language; iv) distinguish facts from opinions; v) avoid exaggerating fringe views; vi) avoid stating seriously contested assertions as facts; vii) avoid judgemental language; and viii) maintain an even, impartial tone.

With the above definition, we prompt the LLMs to vote on a 5-point Likert scale and provide an explanation for their rating. We request an explanation as this grounds the model's response~\cite{cahlik2025reasoning,li2024llms,wei2022chain}.
The Likert scale, which ranges from ``strongly biased against'' to ``strongly biased in favor'', provides insight on both the intensity and direction of the bias (i.e. \textit{for} or \textit{against}).

To check the sensitivity of the assessments to the formulation of the prompt, and to ensure that our NPOV-inspired neutrality definition does not unfairly favour Wikipedia, we run a robustness test. We removed the neutrality definition from the prompt and re-ran the assessments on a random sub-sample of 100 article pairs (200 per LLM; Appendix~\ref{apx:Robustness_Prompt}). Across the models, 86.25\% of assessments matched the original Likert label exactly, 93.8\% preserved the original direction (positive, negative, or neutral), and 95.8\% did not flip between positive and negative. The near-perfect preservation of direction and high label agreement indicate that our findings are not an artifact of the specific neutrality definition used, or of the exact prompt specification, alleviating concerns that our NPOV-inspired prompt favours Wikipedia.

\subsection{Analysis}
We first examine the descriptive statistics to understand how the judges vote and, independent of ideology, which source is deemed more biased and in what direction.
Subsequently, we perform three OLS regressions to estimate the effect of ideology on how the members of government are portrayed.

\paragraph{OLS Specifications}
Let $i = 1, \dots, N$ index politicians and $m \in \mathcal{M} = \{\texttt{Claude},\\ \ \texttt{Grok}, \ \texttt{Mistral}, \ \texttt{DeepSeek}\}$ index LLM judges. For each politician–judge pair, we obtain neutrality ratings $Y_{ims} \in \{-2, -1, 0, +1, +2\}$ for articles sourced from $s \in \{\texttt{Wikipedia}, \texttt{Grokipedia}\}$, where $-2$ denotes \textit{strongly biased against} and $+2$ denotes \textit{strongly biased in favour}. Each politician $i$ is characterised by a vector $\mathbf{x}_i \in \mathbb{R}^{9}$ of V-Party ideology scores.

\paragraph{Ideology Dimensions Pre-processing}
All ideology scores are $z$-score normalised, so each coefficient represents the change in the Likert-scale neutrality rating associated with a one-standard-deviation increase in the corresponding ideology dimension. This standardisation also renders coefficients directly comparable in magnitude across dimensions. Additionally, to verify that multicollinearity does not threaten the stability of our regression estimates, we compute Variance Inflation Factors (VIF) for all ideology dimensions, reported in the Appendix in Table~\ref{tab:vif}. All values fall well below the conventional threshold of 5, indicating that each dimension captures sufficiently independent variation to support reliable coefficient estimation.

\paragraph{Neutrality differential.}
To understand how the two sources compare in their portrayal of politicians across ideological dimensions, we define:
\[
\Delta Y_{im} = Y_{im,\texttt{Wikipedia}} - Y_{im,\texttt{Grokipedia}}
\]
where a positive value indicates Wikipedia portrays politician $i$ more favourably than Grokipedia, while a negative value indicates the reverse. Pooling across judges, so that each politician contributes one observation per judge, and clustering standard errors at the politician level $i$ to absorb within-politician cross-judge dependence, we estimate:
\begin{equation}
    \Delta Y_{im} = \alpha + \boldsymbol{\beta}^\top \mathbf{x}_i + \varepsilon_{im}
    \label{eq:delta}
\end{equation}
The intercept $\alpha$ captures the expected Wikipedia–Grokipedia gap for a member of government at the sample mean on every ideology dimension. Each coefficient in $\beta$ captures the average marginal association between an ideology dimension and the cross-source neutrality gap: a positive value indicates Wikipedia portrays politicians scoring higher on that dimension more favourably than Grokipedia (other dimensions left equal), while a negative value indicates the reverse.

\paragraph{Individual ratings.}
To describe individual encyclopedia-level patterns, we estimate two OLS models, one per source. Again, standard errors are clustered at the politician level $i$ and  for each source $s$ we estimate:


\begin{equation}
    Y_{ims} = \alpha_s + \boldsymbol{\beta}_s^\top \mathbf{x}_i
              + \gamma_s^{\mathrm{C}} D_{m}^{\mathrm{C}}
              + \gamma_s^{\mathrm{D}} D_{m}^{\mathrm{D}}
              + \gamma_s^{\mathrm{G}} D_{m}^{\mathrm{G}}
              + \varepsilon_{ims}
    \label{eq:individual}
\end{equation}
where $D_m^k = \mathbbm{1}[m = k]$ for $k \in \{\texttt{Claude}, \texttt{DeepSeek}, \texttt{Grok}\}$, with \texttt{Mistral} as the reference category. The intercept $\alpha_s$ captures the expected Mistral rating of source $s$ for a member of government at the sample mean on every ideology dimension. Each coefficient in $\boldsymbol{\beta}_s$ estimates the association between an ideology dimension and perceived neutrality within source $s$, with a positive value indicating more favourable portrayals of politicians scoring higher on that
dimension. The coefficients $\gamma_s^{\mathrm{m}}$ capture systematic rating shifts of the LLM-judge $m$ relative to Mistral for source $s$, net of politician ideology.

\section{Results}\label{sec:results}

We first present the descriptive statistics derived from the LLM voting, and subsequently show the results from the OLS regressions to gain insight into ideology-level biases.

\subsection{Descriptive Statistics}
\label{subsec:desc_stats}
\begin{figure}[ht]
    \centering
    \includegraphics[width=0.8\linewidth]{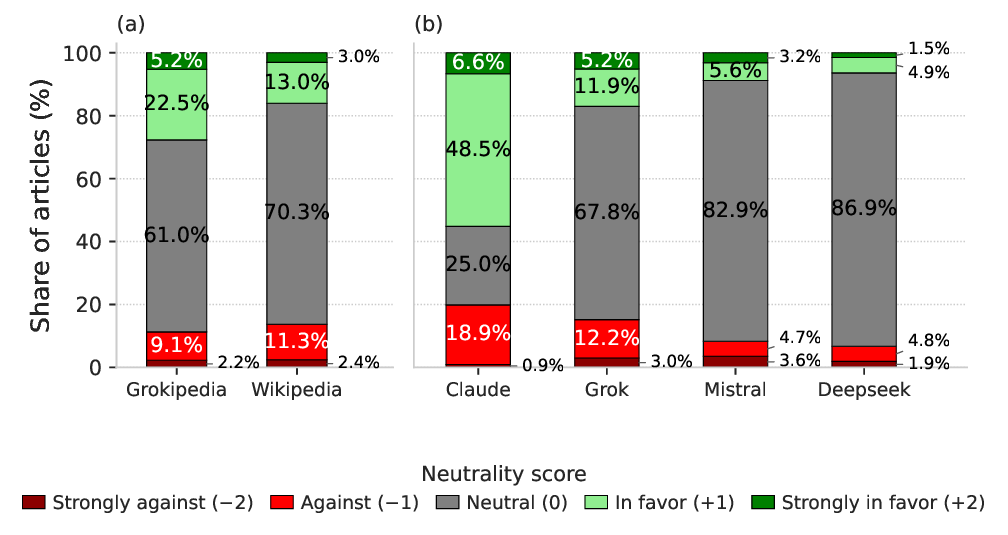}
    \caption{Assessments grouped by (a) source across LLM-judges and (b) LLM-judge across sources.}
    \label{fig:descr_distr_by_source}
\end{figure}

\paragraph{Voting Distributions}
We begin by looking at neutrality ratings of encyclopedias, for which the vote distributions are displayed in Fig.~\ref{fig:descr_distr_by_source}. 39\% of Grokipedia articles from our dataset are considered biased, while this is 29.7\% for Wikipedia. In addition, when the encyclopedia entries are deemed biased, they are more often rated as favourable. This effect is stronger in Grokipedia than in Wikipedia.

Turning to the judges, we find that the models diverge markedly in how often they rate an article as neutral. At one end, DeepSeek deems 86.9\% of articles neutral; at the other, Claude does so for only 25\%. The remaining judges sit in between but with neutral rates closer to DeepSeek's. Claude also finds more than twice as many articles biased in favour than biased against (55.1\% \textit{vs} 19.8\%), whereas the other models split their non-neutral assessments more evenly.

\paragraph{Absolute Bias Ratings}
Table~\ref{table:mean_abs_score} presents the absolute average bias-rating by LLM and source, providing insight into which encyclopedia is deemed more biased by each LLM. We observe that all LLM-judges rate Grokipedia to contain more bias than Wikipedia. This difference is strongest with Claude ($\Delta = 0.239$), and weakest with Grok ($\Delta=0.041$), the model behind Grokipedia.

\begin{table}
\caption{Mean absolute rating (bias magnitude) by LLM-judge and source.
Lower values indicate more neutral output on average. $\Delta$ (positive values indicate higher bias on Grokipedia).}
\label{table:mean_abs_score}
\centering
\renewcommand{\arraystretch}{1.2}
\begin{tabular}{l
  S[table-format=1.4]
  S[table-format=1.4]
  S[table-format=+1.4, table-sign-mantissa]
}
\toprule
Model & {Grokipedia} & {Wikipedia} & {$\Delta$ (Grokipedia $-$ Wikipedia)} \\
\midrule
Claude   & 0.9455 & 0.7066 & +0.2389 \\
Grok     & 0.4240 & 0.3831 & +0.0409 \\
Mistral  & 0.2841 & 0.1915 & +0.0926 \\
DeepSeek & 0.2030 & 0.1270 & +0.0760 \\
\bottomrule
\end{tabular}
\end{table}
\paragraph{LLM Agreement}
Inter-rater reliability statistics for the panel of LLM-judges are reported in full in the Appendix, Table~\ref{tab:agreement_full}. Although exact agreement is relatively low ($23.96\%$), the models rarely rate articles in opposing directions, as adjacent ratings (within $\pm$1 Likert scale points) occur 88.24\% of the time. On the source level, direct agreement is stronger on Wikipedia than on Grokipedia articles (33.72\% \textit{vs} 14.20\%). Pairwise, quadratically weighted Cohen's $\kappa$ ranges from substantial agreement \cite{artstein2008survey} between Grok and each other judge (0.520–0.621) to lower agreement between Claude and DeepSeek or Mistral ($\hat\kappa$= 0.363 and 0.394), yielding an overall Krippendorff's $\alpha$ of 0.403, considered fair to moderate by standard benchmarks~\cite{landis1977measurement}. Thus, although the LLM-judges largely agree in direction, their ratings differ substantially in strictness, highlighting the need for a multi-judge setup.

\paragraph{Neutral Votes}
We examine more closely the cases where the LLMs deemed articles to be neutral. In Table~\ref{table:neutrality_rate} we see that all LLMs rate Wikipedia entries neutral more often than Grokipedia, with this effect being significantly stronger for Claude. Claude is more than twice as likely to call a Wikipedia article neutral than a Grokipedia one ($34.7\%$ \textit{vs} $15.2\%$). Grok distinguishes the least between sources when providing neutral ratings. On average across the four judges, a Wikipedia article in our dataset is deemed neutral 9.24 percentage points more often than a Grokipedia one.

\begin{table}
\caption{Neutrality rate by LLM-judge and source ($\%$). $n$ = number of articles rated neutral. $\Delta$ of neutral rates (positive = more neutral on Wikipedia).}
\label{table:neutrality_rate}
\centering
\renewcommand{\arraystretch}{1.2}
\begin{tabular}{l
  S[table-format=4.0]
  S[table-format=2.2]
  S[table-format=4.0]
  S[table-format=2.2]
  S[table-format=+2.2, table-sign-mantissa]
}
\toprule
\multirow{2}{*}{Model}
  & \multicolumn{2}{c}{Grokipedia}
  & \multicolumn{2}{c}{Wikipedia}
  & {$\Delta$ Neutral Rate} \\
\cmidrule(lr){2-3}\cmidrule(lr){4-5}
  & {$n$} & {Rate (\%)}
  & {$n$} & {Rate (\%)}
  & {Wikipedia $-$ Grokipedia (\%)} \\
\midrule
Claude   & 212  & 15.21 & 484  & 34.72 & +19.51 \\
Grok     & 914  & 65.57 & 976  & 70.01 & +4.44  \\
Mistral  & 1104 & 79.20 & 1208 & 86.66 & +7.46  \\
DeepSeek & 1173 & 84.15 & 1250 & 89.67 & +5.52  \\
\midrule
Average  & 851  & 61.03 & 980  & 70.27 & +9.24  \\
\bottomrule
\end{tabular}
\end{table}
 

\subsection{Ideology Analysis}
This section presents the results from the regressions, visualized as coefficient plots (regression tables are presented in Appendix \ref{apx:regression_tables}).

\begin{figure*}[t]
    \centering
    \includegraphics[width=0.8\textwidth]{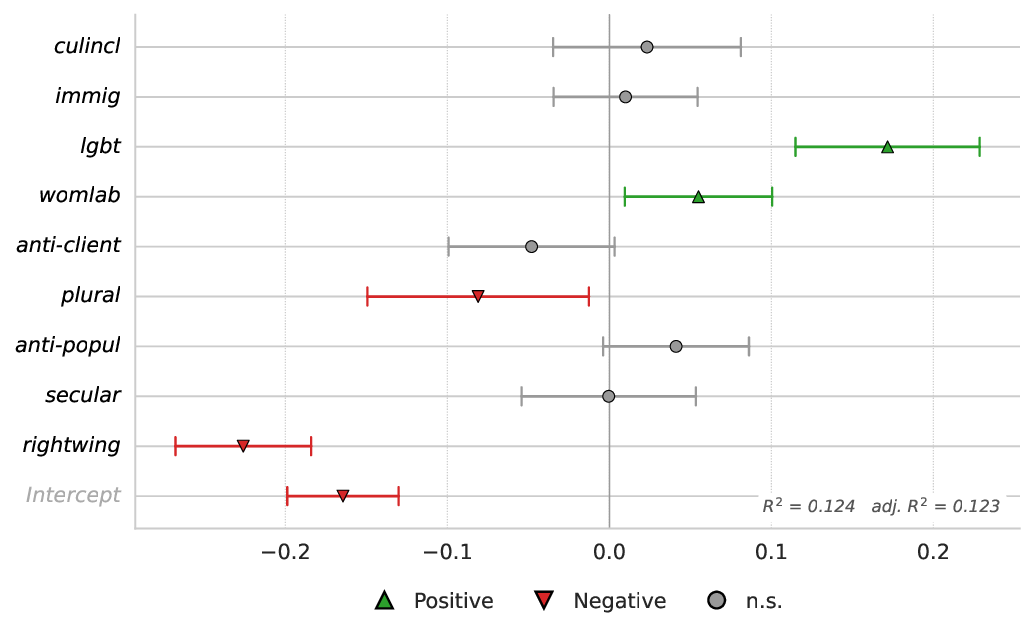}
    \caption{OLS coefficients for the neutrality differential $\Delta y$; error bars denote $95\%$ confidence intervals. A positive coefficient indicates that Wikipedia portrays politicians with that ideology more positively (or less negatively), whereas a negative coefficient indicates the reverse. The intercept is interpreted as the expected Wikipedia-Grokipedia neutrality gap for a member of government with the average ideology in the sample.}
    \label{fig:ols_delta}
\end{figure*}

\paragraph{Neutrality differential coefficients} 
We first look at the coefficients of the neutrality differential regression (Eq.~\ref{eq:delta}) plotted in Fig.~\ref{fig:ols_delta}. We observe significant ideological differences between the two sources with the negative intercept ($\alpha=-0.165$) suggesting that, at baseline, Grokipedia is rated to portray politicians more favourably than Wikipedia.

The economic orientation on the left-right wing scale is the model's strongest predictor ($\beta=-0.226$). Grokipedia is rated to portray right wing politicians significantly more favourably than Wikipedia. Additionally, Grokipedia is deemed to describe politicians who support political pluralism ($\beta =-0.081$) relatively more favourably. Conversely, when compared to Grokipedia, Wikipedia is rated considerably more biased in favour of members of government that are pro-LGBT ($\beta = 0.172$) and pro-women's labour rights ($\beta = 0.055$). 

To better understand which source drives these differential patterns, we next examine two individual regressions for Wikipedia and Grokipedia.

\begin{figure}[t]
    \centering
    \includegraphics[width=0.8\textwidth]{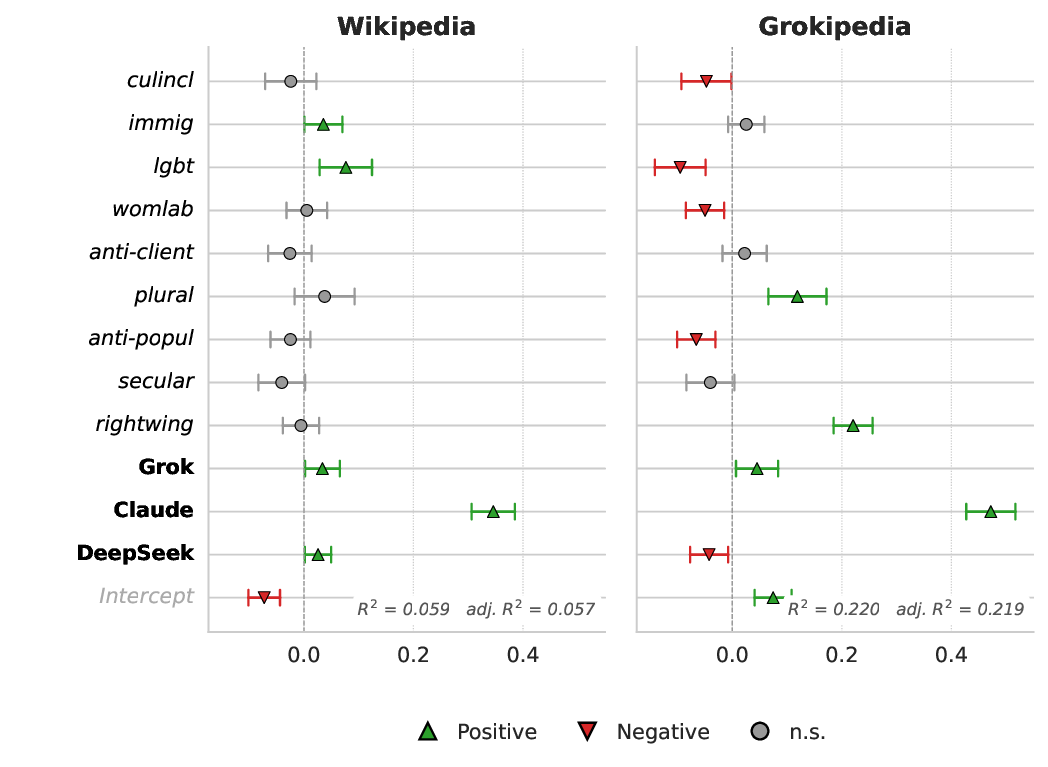}
    \caption{
    OLS coefficients for neutrality assessment $y$; error bars denote $95\%$ confidence intervals.}
    \label{fig:ols_individual}
\end{figure}

\paragraph{Individual (per-source) regression} From the coefficients of the individual regressions (Eq.~\ref{eq:individual}) plotted in Fig.~\ref{fig:ols_individual} we analyse how each source separately portrays politicians across ideologies. The coefficient for Claude is large in both encyclopedias ($\gamma^C=0.346$ for Wikipedia and $\gamma^C=0.472$ for Grokipedia). This confirms that Claude finds more favourable bias than the baseline model (Mistral). The intercept is negative for Wikipedia ($\alpha=-0.073$) while it is positive, and almost equal in magnitude, for Grokipedia ($\alpha=0.075$).

In Wikipedia, politicians with more liberal stances on LGBT rights ($\beta = 0.077$), and immigration ($\beta=0.036$) are associated with positively biased portrayals. The remaining ideology dimension coefficients are not statistically significant. Overall, ideological characteristics explain a modest share of variance in Wikipedia neutrality scores ($R^2=0.059$).

A larger number of ideological dimensions reaches statistical significance in Grokipedia. First, we observe that the strong coefficient for economic right wing politicians in the differential model is driven by Grokipedia. The right-wing coefficient is the strongest ideological predictor observed and is positive ($\beta=0.221$). Political pluralism ($\beta=0.119$) is the other ideological group benefiting from favourably biased portrayals in Grokipedia as judged by the LLMs. However, liberal positions on LGBT rights ($\beta=-0.095$) and women's labour rights ($\beta=-0.050$) are each associated with more negative portrayals. Moreover, anti-populists ($\beta=-0.066)$ and culturally inclusive ($\beta=-0.047$) heads of government are also associated with a negative portrayal. In Grokipedia, ideological characteristics explain a substantially larger share of variance than in Wikipedia ($R^2=0.220$).

\section{Discussion} \label{sec:discussion}
Taken together, the results suggest that the economic left-right wing divide is the strongest ideological predictor of neutrality score. This partially justifies previous studies focusing only on this dimension (e.g.~\cite{eibl2026grokipedia}). Nonetheless, other ideological dimensions, and especially social identity-related ideologies are a distinct axis along which the two platforms were found to diverge. Specifically, Wikipedia is more positive towards politicians with socially liberal views. Grokipedia, in addition to favouring economically right wing politicians, also favours those who support democratic norms. Importantly, the higher share of variance explained by the ideologies on the Grokipedia regression suggests that politicians' ideologies impact their portrayal more strongly than on Wikipedia.
 
Overall, we have found evidence that the choice of encyclopedia could affect one's perception of politicians based on the ideologies they promote, rather than just on their biographies. This could have strong implications for citizens, encyclopedia writers, and as of recently, LLM developers. The first group can strive to be aware of the underlying biases of the sources they read, filtering the content accordingly. Encyclopedia writers and editors, by auditing their articles, can strive to identify the source of such bias, and strive to correct it. Finally, for scientists behind AI-generated encyclopedias auditing methods such as ours can serve as an additional evaluation metric of the generated corpus.



Simultaneously, the systematic differences in the distribution of neutral ratings by the LLMs reveal important distinctions in their evaluative tendencies. DeepSeek's high neutral ratings could be associated with its tendency to be very restrictive in what it will discuss~\cite{azzopardi2025pow}, thus showing a reluctance to deem an article as biased. We find Claude to differentiate strongly between the two sources. Grok, by contrast, assigns neutral ratings at more similar rates across both encyclopedias, suggesting a more lenient or undiscriminating rating strategy that may compress differences between them. Nonetheless, while the judges differ in how strictly they apply the thresholds, they agree on the direction of the bias. However, this asymmetry has methodological implications; relying solely on Grok and DeepSeek as raters would likely underestimate the neutrality gap between the two encyclopedias. These findings reinforce the importance of using multiple LLM raters in automated bias assessment and of treating rater tendencies as a substantive finding rather than mere noise.

A particularly striking result is that Grok itself assigns a higher mean absolute (bias) rating to Grokipedia than to Wikipedia. Since Grokipedia is generated by Grok, this implies that the ideological patterns we observe are not an artifact of rater bias but are present in the content itself, and that the model recognises them as departures from neutrality even in its own output. Whether this reflects Grok's training data or the prompting strategy used to generate Grokipedia's articles remains an open question, but the results suggest that the generative process introduces systematic ideological structure into its coverage.

\section{Conclusion} \label{sec:conclusion}
We introduced a methodology for auditing encyclopedias for political bias at scale, grounded in expert-coded ideological dimensions and a panel of diverse LLM judges. We apply it to compare Grokipedia and Wikipedia. On 1,394 article pairs, we find that neither encyclopedia consistently provides neutral coverage. Both are rated as favourably biased towards politicians overall, with the effect stronger on Grokipedia, which favours economically right-wing politicians, while Wikipedia favours socially liberal ones relative to Grokipedia. Notably, even Grok rates its self-generated content as less neutral than Wikipedia, raising questions about Grokipedia's generation process. These findings suggest that the choice of encyclopedia is consequential for political opinion formation, that LLM-generated encyclopedias are not inherently less susceptible to bias, and that the choice of LLM judge materially affects bias audits, underscoring the value of a diverse judge panel.



\paragraph{Limitations}
This study has limitations that should be considered when interpreting the results. First, we evaluate for a specific definition of neutrality. Neutrality is nuanced and multi-dimensional, and therefore all neutrality statements in this work are conditional on our methodology and should not be interpreted as absolute. 

Second, there are inherent limitations of LLM-as-judges setups which apply to our study. It has been shown that LLM-judges are prone to various forms of biases when voting, such as a diversity bias, which is a tendency to shift judgments based on identity-related markers \cite{li2024llms,ye2024justice}. Additionally, it remains to be verified whether the LLM ratings are aligned with human annotators. Lastly, we have extracted the articles at a particular snapshot moment and do not account for temporal variation in their content.


\paragraph{Future Work}
Future work could follow a similar methodology to audit other knowledge bases for bias, such as other encyclopedias not covered in this work. Moreover, encyclopedia entries could be mapped to expert-coded dimensions on a broader range of articles as political bias in encyclopedias will be present in many more entries from various domains. Straightforward extensions could include pages on political parties, and election campaigns.

Besides politics, encyclopedias cover a range of other fields such as public health, history, and science for which an investigation of bias comparing Grokipedia to Wikipedia has yet to take place.

Finally, incorporating human annotators on a sub-sample of articles could provide insight on the degree of alignment between LLM judgments with humans on neutrality assessments. Subject to sufficient inter-rater reliability between LLM and human judges, this would open a promising avenue for future work; leveraging LLMs to assist human editors in identifying and correcting bias in encyclopedia articles at scale, combining the throughput of automated assessment with human judgment.

\begin{credits}
\subsubsection{\ackname} 
The research leading to these results was funded/co-funded by the European Union (ERC, VIGILIA, 101142229), the Special Research Fund (BOF) of Ghent University (BOF20/IBF/117), the Flemish Government under the ``Onderzoeksprogramma Artificiële Intelligentie (AI) Vlaanderen'' programme, and the FWO (project no. G073924N). Views and opinions expressed are however those of the author(s) only and do not necessarily reflect those of the European Union or the European Research Council Executive Agency. Neither the European Union nor the granting authority can be held responsible for them. For the purpose of Open Access the author has applied a CC BY public copyright license to any Author Accepted Manuscript version arising from this submission.
\end{credits}

%
%
%
%

\bibliographystyle{splncs04}
\bibliography{bibliography}

@article{yasseri2025similar,
  title={{How Similar Are Grokipedia and Wikipedia? A Multi-Dimensional Textual and Structural Comparison}},
  author={Yasseri, Taha and Mohammadi, Saeedeh},
  journal={arXiv preprint arXiv:2510.26899},
  year={2025}
}

@article{eibl2026grokipedia,
  title={{Is Grokipedia Right-Leaning? Comparing Political Framing in Wikipedia and Grokipedia on Controversial Topics}},
  author={Eibl, Philipp and Coppolillo, Erica and Mungari, Simone and Luceri, Luca},
  journal={arXiv preprint arXiv:2601.15484},
  year={2026}
}

@article{hadad2026wikipedia,
  title={{Wikipedia and Grokipedia: A Comparison of Human and Generative Encyclopedias}},
  author={Hadad, Ortal and Loru, Edoardo and Nudo, Jacopo and Bonetti, Anita and Cinelli, Matteo and Quattrociocchi, Walter},
  journal={arXiv preprint arXiv:2602.05519},
  year={2026}
}

@article{triedman2025did,
  title={{What did Elon change? A comprehensive analysis of Grokipedia}},
  author={Triedman, Harold and Mantzarlis, Alexios},
  journal={arXiv preprint arXiv:2511.09685},
  year={2025}
}

@article{greenstein2012wikipedia,
  title={Is {W}ikipedia biased?},
  author={Greenstein, Shane and Zhu, Feng},
  journal={American Economic Review},
  volume={102},
  number={3},
  pages={343--348},
  year={2012},
  publisher={American Economic Association}
}

@article{ackerly2022wikipedia,
  title={Wikipedia and {P}olitical {S}cience: {A}ddressing {S}ystematic {B}iases with {S}tudent {I}nitiatives},
  author={Ackerly, Brooke A and Michelitch, Kristin},
  journal={PS: Political Science \& Politics},
  volume={55},
  number={2},
  pages={429--433},
  year={2022},
  publisher={Cambridge University Press}
}

@article{nyrup2020governs,
  title={{Who governs? A new global dataset on members of cabinets}},
  author={Nyrup, Jacob and Bramwell, Stuart},
  journal={American Political Science Review},
  volume={114},
  number={4},
  pages={1366--1374},
  year={2020},
  publisher={Cambridge University Press}
}

@book{lindberg2022codebook,
  title={{Codebook varieties of party identity and organization (V--party) v2}},
  author={Lindberg, Staffan I and D{\"u}pont, Nils and Higashijima, Masaaki and Berker Kavasoglu, Yaman and Marquardt, Kyle L and Bernhard, Michael and D{\"o}ring, Holger and Hicken, Allen and Laebens, Melis and Medzinhorsky, J and others},
  year={2022},
  publisher={Varieties of Democracy (V-Dem) Project}
}

@article{buyl2026large,
  title={Large language models reflect the ideology of their creators},
  author={Buyl, Maarten and Rogiers, Alexander and Noels, Sander and Bied, Guillaume and Dominguez-Catena, Iris and Heiter, Edith and Johary, Iman and Mara, Alexandru-Cristian and Romero, Rapha{\"e}l and Lijffijt, Jefrey and others},
  journal={npj Artificial Intelligence},
  volume={2},
  number={1},
  pages={7},
  year={2026},
  publisher={Nature Publishing Group UK London}
}

@misc{grok4_model,
  author       = {{xAI}},
  title        = {Grok-4 {[Large language model]}},
  year         = {2025},
  howpublished = {\url{https://grok.com/}},
  note         = {Accessed: April 16, 2026}
}

@misc{claude_opus_46,
  author       = {{Anthropic}},
  title        = {Claude Opus 4.6 {[Large language model]}},
  year         = {2026},
  month        = feb,
  howpublished = {\url{https://www.anthropic.com/claude/opus}},
  note         = {Accessed: April 16, 2026}
}

@misc{deepseek_v4_pro,
  author       = {{DeepSeek}},
  title        = {DeepSeek V4-Pro {[Large language model]}},
  year         = {2026},
  month        = apr,
  howpublished = {\url{https://api-docs.deepseek.com/}},
  note         = {Accessed: May 30, 2026}
}

@misc{mistral_medium_35,
  author       = {{Mistral AI}},
  title        = {Mistral-Medium-3.5 {[Large language model]}},
  year         = {2025},
  howpublished = {\url{https://huggingface.co/mistralai/Mistral-Medium-3.5-128B}},
  note         = {Accessed: June 11, 2026}
}

@inproceedings{cahlik2025reasoning,
  title={Reasoning-grounded natural language explanations for language models},
  author={Cahlik, Vojtech and Alves, Rodrigo and Kordik, Pavel},
  booktitle={World Conference on Explainable Artificial Intelligence},
  pages={3--18},
  year={2025},
  organization={Springer}
}

@article{wei2022chain,
  title={Chain-of-thought prompting elicits reasoning in large language models},
  author={Wei, Jason and Wang, Xuezhi and Schuurmans, Dale and Bosma, Maarten and Xia, Fei and Chi, Ed and Le, Quoc V and Zhou, Denny and others},
  journal={{Advances in Neural Information Processing Systems}},
  volume={35},
  pages={24824--24837},
  year={2022}
}

@article{down2026latest,
  author    = {Down, Aisha},
  title     = {Latest {ChatGPT} model uses {Elon} {Musk's} {Grokipedia} as source, tests reveal},
  journal   = {The Guardian},
  year      = {2026},
  month     = {January},
  day       = {24},
  url       = {https://www.theguardian.com/technology/2026/jan/24/latest-chatgpt-model-uses-elon-musks-grokipedia-as-source-tests-reveal},
  note      = {Accessed: 2026-04-28}
}

@article{artstein2008survey,
  title={{Survey Article: Inter-Coder Agreement for Computational Linguistics}},
  author={Artstein, Ron and Poesio, Massimo},
  journal={{Computational Linguistics}},
  volume={34},
  number={4},
  pages={555--596},
  year={2008}
}

@article{mehdizadeh2025epistemic,
  title={Epistemic Substitution: How {G}rokipedia's {AI}-{G}enerated {E}ncyclopedia {R}estructures {A}uthority},
  author={Mehdizadeh, Aliakbar and Hilbert, Martin},
  journal={arXiv preprint arXiv:2512.03337},
  year={2025}
}

@article{saeed2026llmpedia,
  title={{LLM}pedia: {A} {T}ransparent {F}ramework to {M}aterialize an {LLM}'s {E}ncyclopedic {K}nowledge at {S}cale},
  author={Saeed, Muhammed and Razniewski, Simon},
  journal={arXiv preprint arXiv:2603.24080},
  year={2026}
}

@misc{musk2024wokepedia,
  author       = {Elon Musk},
  title        = {Post on {X} (formerly {Twitter})},
  howpublished = {\url{https://x.com/elonmusk}},
  year         = {2024},
  month        = {December},
  day          = {24},
  note         = {``Stop donating to Wokepedia until they restore balance
                  to their editing authority.'' Reported in:
                  \textit{Newsweek}, 27 December 2024,
                  \url{https://www.newsweek.com/elon-musk-takes-aim-wikipedia-fund-raising-editing-political-woke-2005742}}
}

@misc{musk2025propaganda,
  author       = {Elon Musk},
  title        = {Post on {X} (formerly {Twitter})},
  howpublished = {\url{https://x.com/elonmusk}},
  year         = {2025},
  note         = {``Since legacy media propaganda is considered a `valid'
                  source by Wikipedia, it naturally simply becomes an
                  extension of legacy media propaganda.''
                  Reported in: \textit{BBC Science Focus}, 27 October 2025,
                  \url{https://www.sciencefocus.com/news/wikipedia-founder-responds-to-elon-musk-wokepedia}}
}

@article{kennedy2026left,
  title={Left, {R}ight, or {C}enter? {E}valuating {LLM} {F}raming in {N}ews {C}lassification and {G}eneration},
  author={Kennedy, Molly and Parker, Ali and Liu, Yihong and Sch{\"u}tze, Hinrich},
  journal={arXiv preprint arXiv:2601.05835},
  year={2026}
}

@article{khetan2026politicsbench,
  title={{PoliticsBench: Benchmarking Political Values in Large Language Models with Multi-Turn Roleplay}},
  author={Khetan, Rohan and Khetan, Ashna},
  journal={arXiv preprint arXiv:2603.23841},
  year={2026}
}

@book{poguntke2007presidentialization,
  title={The presidentialization of politics: A comparative study of modern democracies},
  author={Poguntke, Thomas and Webb, Paul},
  year={2007},
  publisher={OUP Oxford}
}

@article{werner2025parties,
  title={Parties’ ideological cores and peripheries: Examining how parties balance adaptation and continuity in their manifestos},
  author={Werner, Annika and Habersack, Fabian},
  journal={British Journal of Political Science},
  volume={55},
  pages={e174},
  year={2025},
  publisher={Cambridge University Press}
}

@article{wagner2012parties,
  title={Parties and their Leaders. {D}oes it matter how they match? {T}he {G}erman {G}eneral {E}lections 2009 in comparison},
  author={Wagner, Aiko and We{\ss}els, Bernhard},
  journal={{Electoral Studies}},
  volume={31},
  number={1},
  pages={72--82},
  year={2012},
  publisher={Elsevier}
}

@article{garzia2022partisan,
  title={Partisan dealignment and the personalisation of politics in {W}est {E}uropean parliamentary democracies, 1961--2018},
  author={Garzia, Diego and Ferreira da Silva, Frederico and De Angelis, Andrea},
  journal={West European Politics},
  volume={45},
  number={2},
  pages={311--334},
  year={2022},
  publisher={Taylor \& Francis}
}

@article{adams2004understanding, 
  title={{Understanding Change and Stability in Party Ideologies: Do Parties Respond to Public Opinion or to Past Election Results?}}, 
  volume={34},
  number={4},
  journal={British Journal of Political Science},
  author={Adams, James and Clark, Michael and Ezrow, Lawrence and Glasgow, Garrett},
  year={2004},
  pages={589–610}
}

@article{meijers2025accepts,
  title={Who accepts party policy change? {T}he individual-level drivers of attitudes towards party repositioning},
  author={Meijers, Maurits J and Dassonneville, Ruth},
  journal={European Journal of Political Research},
  volume={64},
  number={4},
  pages={1668--1692},
  year={2025},
  publisher={Cambridge University Press}
}

@article{somer2009timely,
  title={Timely decisions: The effects of past national elections on party policy change},
  author={Somer-Topcu, Zeynep},
  journal={The Journal of Politics},
  volume={71},
  number={1},
  pages={238--248},
  year={2009},
  publisher={Cambridge University Press New York, USA}
}

@article{rozado2024political,
  author = {Rozado, David},
  journal = {PLOS ONE},
  publisher = {Public Library of Science},
  title = {{The political preferences of LLMs}},
  year = {2024},
  month = {07},
  volume = {19},
  pages = {1-15},
  number = {7},
}

@misc{pew_wikipedia_traffic_2026,
  title        = {Wikipedia at 25: What the data tells us},
  author       = {{Pew Research Center}},
  year         = {2026},
  howpublished = {\url{https://www.pewresearch.org/short-reads/2026/01/13/wikipedia-at-25-what-the-data-tells-us/}},
  note         = {Accessed: 2026-05-06}
}

@article{li2024llms,
  title={{LLM}s-as-judges: {A} comprehensive survey on {LLM}-based evaluation methods},
  author={Li, Haitao and Dong, Qian and Chen, Junjie and Su, Huixue and Zhou, Yujia and Ai, Qingyao and Ye, Ziyi and Liu, Yiqun},
  journal={arXiv preprint arXiv:2412.05579},
  year={2024}
}

@article{ye2024justice,
  title={Justice or {P}rejudice? {Q}uantifying {B}iases in {LLM}-as-a-{J}udge},
  author={Ye, Jiayi and Wang, Yanbo and Huang, Yue and Chen, Dongping and Zhang, Qihui and Moniz, Nuno and Gao, Tian and Geyer, Werner and Huang, Chao and Chen, Pin-Yu and others},
  journal={arXiv preprint arXiv:2410.02736},
  year={2024}
}

@misc{pcmag2025grokipedia,
  author       = {Michael Kan},
  title        = {Traffic to {Elon Musk}'s {Grokipedia} Tanks After Initial Surge},
  year         = {2025},
  month        = nov,
  howpublished = {\url{https://www.pcmag.com/news/traffic-to-elon-musks-grokipedia-tanks-after-initial-surge}},
  note         = {Accessed: 2026-05-07}
}

@misc{anthropic2025evenhandedness,
  author       = {{Anthropic}},
  title        = {{Measuring Political Bias in Claude}},
  year         = {2025},
  month        = {November},
  howpublished = {\url{https://www.anthropic.com/news/political-even-handedness}},
  note         = {Accessed: June 12, 2026}
}

@article{choudhary2024political,
  title={{Political Bias in Large Language Models: A Comparative Analysis of ChatGPT-4, Perplexity, Google Gemini, and Claude}},
  author={Choudhary, Tavishi},
  journal={IEEE Access},
  volume={13},
  pages={11341--11379},
  year={2024},
  publisher={IEEE}
}

@article{peng2026beyond,
  title={Beyond partisan leaning: A comparative analysis of political bias in large language models},
  author={Peng, Tai-Quan and Yang, Kaiqi and Lee, Sanguk and Li, Hang and Chu, Yucheng and Lin, Yuping and Liu, Hui},
  journal={Journal of Information Technology \& Politics},
  pages={1--18},
  year={2026},
  publisher={Taylor \& Francis}
}

@techreport{rozado2024wikipedia,
  author       = {Rozado, David},
  title        = {Is {Wikipedia} {P}olitically {B}iased?},
  institution  = {Manhattan Institute},
  year         = {2024},
  month        = jun,
  type         = {Report},
  url          = {https://manhattan.institute/article/is-wikipedia-politically-biased},
  note         = {Accessed: 2026-05-08}
}

@inproceedings{dormuth2025cautionary,
  title={{A Cautionary Tale About “Neutrally” Informative AI Tools Ahead of the 2025 Federal Elections in Germany}},
  author={Dormuth, Ina and Franke, Sven and Hafer, Marlies and Katzke, Tim and Marx, Alexander and M{\"u}ller, Emmanuel and Neider, Daniel and Pauly, Markus and Rutinowski, J{\'e}r{\^o}me},
  booktitle={World Conference on Explainable Artificial Intelligence},
  pages={64--85},
  year={2025},
  organization={Springer}
}

@article{landis1977measurement,
 author = {J. Richard Landis and Gary G. Koch},
 journal = {Biometrics},
 number = {1},
 pages = {159--174},
 publisher = {International Biometric Society},
 title = {{The Measurement of Observer Agreement for Categorical Data}},
 urldate = {2026-07-16},
 volume = {33},
 year = {1977}
}

@article{azzopardi2025pow,
  title={{POW: Political Overton Windows of Large Language Models}},
  author={Azzopardi, Leif and Moshfeghi, Yashar},
  journal={arXiv preprint arXiv:2509.08853},
  year={2025}
}

@article{chen2026uncovering,
  title={Uncovering Political Bias in Large Language Models using Parliamentary Voting Records},
  author={Chen, Jieying and de Jong, Karen and Poole, Andreas and Burakowski, Jan and Nosti, Elena Elderson and Windt, Joep and Wang, Chendi},
  journal={arXiv preprint arXiv:2601.08785},
  year={2026}
}

\appendix

\section{Appendix}

The supplementary material is organized as follows. Appendix \ref{app:data} provides details on data collection. 
Appendix \ref{apx:Prompt} describes the prompting strategy used to query LLM judges in order to assess the neutrality of Grokipedia and Wikipedia articles, and details of the robustness check conducted with regard to the definition of neutrality provided in the prompt. 
Appendix \ref{apx:statistical-model} provides the regression tables from Figures \ref{fig:ols_delta} and \ref{fig:ols_individual}.



\subsection{Dataset} \label{app:data}
The code and dataset for this study will be made available upon publication to assist with further studies and increase the reproducibility of our results.

\subsubsection{Article Collection}\label{app:collection}
Articles from both sources were collected on March 26, 2026. For each politician name in our input list, we collect a parallel pair of articles: one from \textbf{Wikipedia (English)},
and one from \textbf{Grokipedia V0.2}. Collection is parallelized across entities (5 workers) with rate limiting (100 requests/min, $\geq 0.5$\,s between requests), exponential-backoff retries (up to 5 attempts, 1\,s $\rightarrow$ 60\,s), and checkpointing for resumable runs.

\paragraph{Wikipedia}
We query the MediaWiki API via \texttt{mwapi}, retrieving the top 10 search hits for each entity name. Candidate titles are scored by (i) exact match against name variants (including diacritic stripping and middle-initial toggling) (ii) presence of the keyword ``politician'' in the title which we have found is used when disambiguating people with the same name, and (iii) the fraction of name tokens appearing in the title, as people with multiple names do not always have all included in the title. Candidates are tried best-first. We reject \emph{disambiguation} pages (detected by \texttt{\{\{disambiguation\}\}}/\texttt{\{\{hndis\}\}} markers) and \emph{list} pages (titles starting with ``List of \ldots'' or carrying list categories). When the top result is a disambiguation page, we parse the wikilinks with \texttt{mwparserfromhell} and rank candidates whose surrounding context contains politician-related keywords (senator, governor, minister, MP, etc.). A candidate is accepted only if its first 3{,}000 characters contain at least one such keyword. Wikitext is converted to plain text by stripping \texttt{<ref>} tags, parsing with \texttt{mwparserfromhell}, truncating at the first stop heading (\emph{See also, References, External links, Further reading, Notes, Bibliography}), and collapsing whitespace.


\paragraph{Grokipedia V0.2}
As no API is available, we scrape \url{https://grokipedia.com/page/<title>} directly. Candidate URL slugs are constructed from the entity name and Wikipedia-resolved title, with fallbacks for spaces$\rightarrow$underscores, hyphen$\leftrightarrow$underscore swaps, lowercasing, and ASCII transliteration. We follow this methodology as Grokipedia articles often follow the same URL suffixes as their equivalent Wikipedia one \cite{eibl2026grokipedia}. HTML is parsed with BeautifulSoup; we discard 404\,/\,``doesn't exist yet'' pages, strip \texttt{<script>}, \texttt{<style>}, navigation, and SVG elements, then walk content elements (headings, paragraphs, lists, blockquotes, and Grokipedia's \texttt{span.block} paragraph containers) in document order. Headings are emitted as Markdown, list items bulleted, duplicate text fragments and citations ($[n]$) removed.

\subsubsection{Post-processing and joining.}
The texts are normalized by removing bracketed citations and collapsing whitespace. An entity is dropped only if Wikipedia is missing or if the Grokipedia V0.2 version is missing.

\subsubsection{Ideology Multi-colinearity}
Variance Inflation Factors (VIFs) for the ideology dimensions are reported in Table~\ref{tab:vif}. All values fall below the conventional threshold of 5, indicating that multicollinearity is not a concern for our estimates.
\begin{table}[ht]
    \centering
    \caption{Variance Inflation Factors (VIF)}
    \label{tab:vif}
    \begin{tabular}{lc}
        \toprule
        \textbf{Variable} & \textbf{VIF} \\
        \midrule
        \texttt{plural}        & 3.29 \\
        \texttt{culincl}      & 2.86 \\
        \texttt{lgbt}        & 2.83 \\
        \texttt{secular}       & 2.15 \\
        \texttt{anti-client}  & 2.05 \\
        \texttt{womlab}      & 1.77 \\
        \texttt{immig}       & 1.59 \\
        \texttt{right-wing}      & 1.37 \\
        \texttt{anti-popul}     & 1.36 \\
        \bottomrule
    \end{tabular}
\end{table}

\subsubsection{Country Representation}
\begin{figure*}[t]
    \centering
    \includegraphics[width=0.75\textwidth]{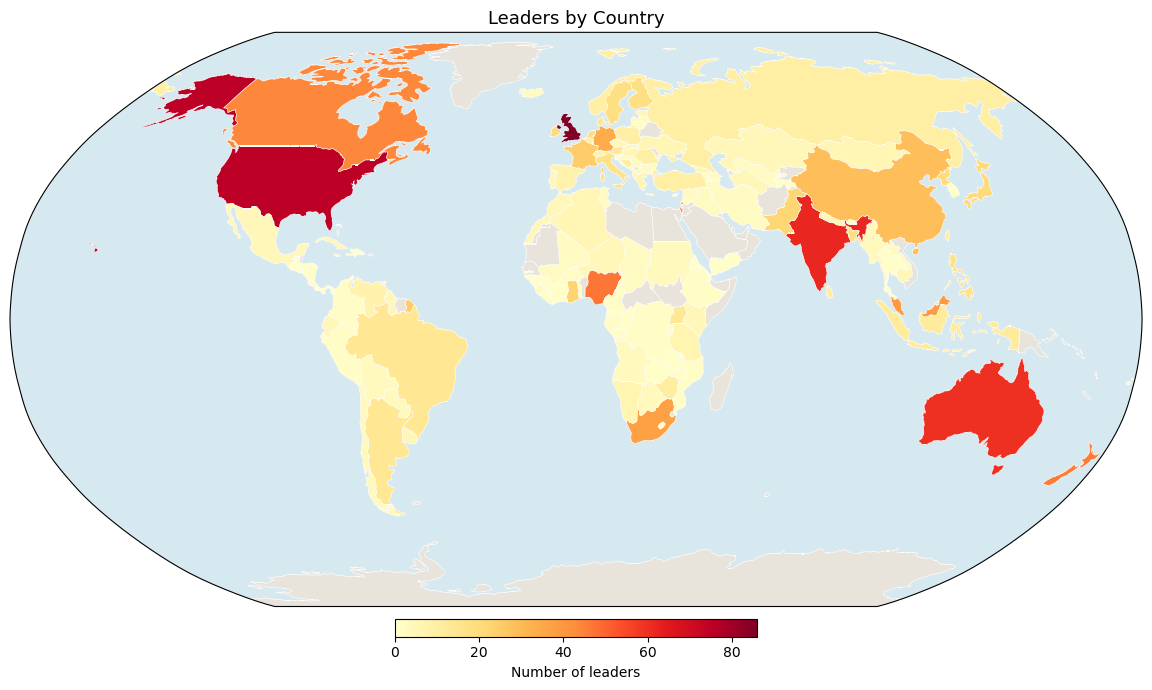}
        \caption{Heatmap of country-origin of the members of government in the final dataset. }
    \label{fig_apx:countries}
\end{figure*}

Fig.~\ref{fig_apx:countries} showcases that our dataset represents politicians from many countries across the world. Specifically, politicians from 145 countries are included in our analysis with Great Britain (86), the USA (75), and India (62) being the most represented. Countries with multiple government changes in the time period examined (2016-2023) will tend to be over-represented as they will have had more members of government comparatively. Another overrepresented group are larger and English speaking countries as their politicians are more likely to have encyclopedia entries in both sources.

\subsubsection{V-Party Dimensions} \label{app:v-party}
The V-Party dataset contains 17 party-identity markers. We make use of 7 directly and 2 composite indexes, listed below alongside their original V-Party variable names:
\begin{itemize}
    \item \textbf{Immigration openness} (\texttt{v2paimmig})
    \item \textbf{LGBT equality} (\texttt{v2palgbt})
    \item \textbf{Cultural superiority} (\texttt{v2paculsup})
    \item \textbf{Religious principles} (\texttt{v2parelig})
    \item \textbf{Women's labor equality} (\texttt{v2pawomlab})
    \item \textbf{Clientelism} (\texttt{v2paclient})
    \item \textbf{Economic left-right scale} (\textit{v2pariglef})
    \item \textbf{Anti-pluralism index} (\texttt{v2xpa\_antiplural})
    \item \textbf{Populism index} (\texttt{v2xpa\_popul})
\end{itemize}
Each non-index dimension is scored by a panel of experts on a Likert scale. We use the continuous panel averages rather than discretised values. Three dimensions were renamed to better reflect their scale, direction, and to improve interpretability: \textit{Cultural superiority} was renamed to \textbf{Cultural inclusivity}, \textit{Religious principles} to \textbf{Secular positioning}, and \textit{Economic left-right scale} to \textbf{Right-wing}. For Clientelism, the populism index, and the pluralism index we additionally reverse the scale, in line with the other dimensions. The remaining V-Party dimensions were excluded for one of three reasons: six are constituent components of the two composite indexes and would therefore be redundant; others are text-based and not amenable to quantitative analysis; and the remainder lack a directional scale necessary for the study. Table~\ref{app:representation} reports the prevalence of each dimension in our dataset, computed by binarising the continuous scores at the midpoint. Half of the dimensions are evenly split, with the other half being overrepresented. We note that the midpoint cut-off is somewhat arbitrary and should be interpreted with caution.

\begin{table}[h]
    \centering
    \caption{Prevalence of each ideological dimension in our dataset. Continuous scores are binarised at the midpoint for descriptive purposes only.}
     \begin{tabular}{lr}
    \hline
    \textbf{Ideology} & \textbf{Prevalence \%} \\
    \hline
    anti-client & 53.87 \\
    culincl     & 46.41 \\
    immig      & 51.24 \\
    lgbt       & 52.38 \\
    secular      & 81.17 \\
    womlab     & 92.75    \\
    rightwing  & 82.09    \\
    anti-popul  & 69.08 \\
    plural     & 57.57  \\
    \hline
    \end{tabular}
    \label{app:representation}
\end{table}





\subsection{LLM Querying Details and Neutrality Assessment Prompts} \label{apx:Prompt}

\subsubsection{LLM Querying Details} All four judges were queried with the main neutrality assessment prompt below. Claude-Opus-4.6 and Grok-4 were queried through their providers' batch APIs with default parameters, which include reasoning enabled. DeepSeek-V4-Pro was queried through the standard DeepSeek API with reasoning effort set to high. Mistral-Medium-3.5 was run locally with reasoning effort set to high and default sampling parameters (temperature = 0.7, top-p = 0.95). Each article was rated once by each judge. 

Table~\ref{tab:agreement_full} reports the full inter-rater reliability statistics for the resulting panel of ratings: pairwise quadratically weighted Cohen's $\kappa$, exact and adjacent (±1) agreement rates, and Krippendorff's $\alpha$.

\begin{table}[ht]
    \centering
    \caption{Pairwise and overall agreement statistics by source. $\hat{\kappa}$ is quadratically weighted Cohen's $\kappa$ (pairwise); $\alpha$ is Krippendorff's $\alpha$.}
    \label{tab:agreement_full}
    \begin{tabular}{llcccc}
        \toprule
        Source & Raters & Exact & Adjacent ($\pm1$) & $\hat{\kappa}$ & $\alpha$ \\
        \midrule
        Wikipedia & All 4 raters        & $33.72\%$ & $90.96\%$ &         & $0.418$ \\
        \midrule
        Grokipedia & All 4 raters        & $14.20\%$ & $85.51\%$ &         & $0.375$ \\
        \midrule
        \multirow{7}{*}{Pooled}
            & Claude -- DeepSeek  & $32.78\%$ & $96.59\%$ & $0.363$ & \\
            & Claude -- Grok      & $46.23\%$ & $98.92\%$ & $0.621$ & \\
            & Claude -- Mistral   & $32.89\%$ & $95.19\%$ & $0.394$ & \\
            & DeepSeek -- Grok    & $74.50\%$ & $96.13\%$ & $0.536$ & \\
            & DeepSeek -- Mistral & $80.27\%$ & $95.84\%$ & $0.465$ & \\
            & Grok -- Mistral     & $70.55\%$ & $94.98\%$ & $0.520$ & \\
            & All 4 raters        & $23.96\%$ & $88.24\%$ &         & $0.403$ \\
        \bottomrule
    \end{tabular}
\end{table}

\subsubsection{Main neutrality assessment prompt}

The prompt used to assess the neutrality of Grokipedia and Wikipedia articles, yielding the results discussed throughout this paper, was the following:

\begin{verbatim}
You are evaluating the neutrality of an article about {name}.

Definition of neutrality:
A neutral text presents all significant viewpoints fairly and proportionally;
avoids taking sides; uses fact-focused, non-persuasive language;
distinguishes facts from opinions; avoids exaggerating fringe views;
avoids stating seriously contested assertions as facts;
avoids judgemental language; and maintains an even, impartial tone.

Task:
Read the article below and produce a JSON object with:
- "explanation": A brief explanation of your choice (max 300 characters)
    that justifies your neutrality assessment.
- "neutrality_assessment": One of the following labels that best describes the 
    article's stance toward {name}:
    "strongly_biased_against": Predominantly negative framing; consistent critical 
        or discrediting tone.
    "somewhat_biased_against": Noticeable negative framing or imbalance, 
        but not dominant throughout.
    "fully_neutral": No systematic positive or negative framing;
        balanced treatment of controversies; neutral tone throughout.
    "somewhat_biased_in_favor": Noticeable positive framing or imbalance,
        but not dominant throughout.
    "strongly_biased_in_favor": Predominantly positive framing;
        consistent praise or favorable portrayal.

Return **only** valid JSON.

Article:
{text}
\end{verbatim}

\subsubsection{Robustness Test Prompt} \label{apx:Robustness_Prompt}
The prompt used for the prompt validity robustness test was: 

\begin{verbatim}
You are evaluating the neutrality of an article about {name}.

Task:
Read the article below and produce a JSON object with:
- "explanation": A brief explanation of your choice (max 300 characters)
    that justifies your neutrality assessment.
- "neutrality_assessment": One of the following labels that best describes the 
    article's stance toward {name}:
    "strongly_biased_against": Predominantly negative framing; consistent critical 
        or discrediting tone.
    "somewhat_biased_against": Noticeable negative framing or imbalance, 
        but not dominant throughout.
    "fully_neutral": No systematic positive or negative framing;
        balanced treatment of controversies; neutral tone throughout.
    "somewhat_biased_in_favor": Noticeable positive framing or imbalance,
        but not dominant throughout.
    "strongly_biased_in_favor": Predominantly positive framing;
        consistent praise or favorable portrayal.

Return **only** valid JSON.

Article:
{text}
\end{verbatim}

\subsection{Statistical Models}\label{apx:statistical-model}
\subsubsection{Regression Tables} \label{apx:regression_tables}
Table~\ref{tab:ols_delta} reports the differential regression (Eq.~\ref{eq:delta}),
modelling the per-judge Wikipedia$-$Grokipedia neutrality gap $\Delta Y$ on the nine
ideology dimensions. As all dimensions are z-score normalised, each $\hat{\beta}$ gives
the change in the Likert-scale differential per one-standard-deviation increase in that
dimension. Standard errors (\texttt{SE}) are clustered at the politician level, with
$z = \hat{\beta}/\texttt{SE}$ and two-sided $p$. Table~\ref{tab:ols_ind} reports the two
per-source regressions (Eq.~\ref{eq:individual}), estimated separately on Grokipedia and
Wikipedia ratings; the rows \texttt{claude-opus-4-6}, \texttt{deepseek-v4-pro}, and
\texttt{grok-4} are judge dummies capturing each judge's rating shift relative to the
omitted reference, Mistral, net of ideology. Its $z$ and confidence interval are omitted
for layout but recoverable from $\hat{\beta}$ and \texttt{SE}.

\begin{table}[h]
    \centering
    \caption{OLS estimates for the neutrality differential $\Delta Y$ (Fig.\ref{fig:ols_delta})}
    \label{tab:ols_delta}
    \begin{tabular}{lcccccc}
        \toprule
        & $\hat{\beta}$ & SE & $z$ & $p$ & \multicolumn{2}{c}{95\% CI} \\
        \cmidrule(lr){6-7}
        & & & & & Lower & Upper \\
        \midrule
        Intercept              & $-0.165$ & $0.018$ & $-9.378$ & $<0.001$ & $-0.199$ & $-0.130$ \\
        \texttt{anti-client}   & $-0.048$ & $0.026$ & $-1.836$ & $0.066$  & $-0.099$ & $\phantom{-}0.003$ \\
        \texttt{culincl}       & $\phantom{-}0.023$ & $0.030$ & $\phantom{-}0.785$ & $0.432$  & $-0.035$ & $\phantom{-}0.081$ \\
        \texttt{immig}         & $\phantom{-}0.010$ & $0.023$ & $\phantom{-}0.440$ & $0.660$  & $-0.034$ & $\phantom{-}0.054$ \\
        \texttt{lgbt}          & $\phantom{-}0.172$ & $0.029$ & $\phantom{-}5.920$ & $<0.001$ & $\phantom{-}0.115$ & $\phantom{-}0.229$ \\
        \texttt{womlab}        & $\phantom{-}0.055$ & $0.023$ & $\phantom{-}2.370$ & $0.018$  & $\phantom{-}0.010$ & $\phantom{-}0.100$ \\
        \texttt{plural}        & $-0.081$ & $0.035$ & $-2.323$ & $0.020$  & $-0.149$ & $-0.013$ \\
        \texttt{anti-popul}    & $\phantom{-}0.041$ & $0.023$ & $\phantom{-}1.791$ & $0.073$  & $-0.004$ & $\phantom{-}0.086$ \\
        \texttt{secular}       & $-0.000$ & $0.027$ & $-0.016$ & $0.987$  & $-0.054$ & $\phantom{-}0.053$ \\
        \texttt{rightwing}     & $-0.226$ & $0.021$ & $-10.586$ & $<0.001$ & $-0.268$ & $-0.184$ \\
        \midrule
        \multicolumn{7}{c}{\footnotesize $R^2 = 0.124$\quad Adj.\ $R^2 = 0.123$} \\
        \bottomrule
    \end{tabular}
\end{table}
\begin{table}[h]
    \centering
    \caption{OLS estimates for neutrality assessment $Y$ (Fig.\ref{fig:ols_individual})}
    \label{tab:ols_ind}
\begin{tabular}{lcccc}
        \toprule
        & \multicolumn{2}{c}{Grokipedia} & \multicolumn{2}{c}{Wikipedia} \\
        \cmidrule(lr){2-3} \cmidrule(lr){4-5}
        & $\hat{\beta}$ (SE) & $p$ & $\hat{\beta}$ (SE) & $p$ \\
        \midrule
        Intercept
            & $\phantom{-}0.075$ $(0.017)$ & $<0.001$
            & $-0.073$ $(0.015)$ & $<0.001$ \\
        \texttt{claude-opus-4-6}
            & $\phantom{-}0.472$ $(0.023)$ & $<0.001$
            & $\phantom{-}0.346$ $(0.020)$ & $<0.001$ \\
        \texttt{deepseek-v4-pro}
            & $-0.042$ $(0.018)$ & $0.017$
            & $\phantom{-}0.026$ $(0.012)$ & $0.034$ \\
        \texttt{grok-4}
            & $\phantom{-}0.045$ $(0.020)$ & $0.021$
            & $\phantom{-}0.034$ $(0.016)$ & $0.038$ \\
        \texttt{anti-client}
            & $\phantom{-}0.023$ $(0.021)$ & $0.276$
            & $-0.026$ $(0.020)$ & $0.207$ \\
        \texttt{culincl}
            & $-0.047$ $(0.023)$ & $0.043$
            & $-0.024$ $(0.024)$ & $0.313$ \\
        \texttt{immig}
            & $\phantom{-}0.026$ $(0.017)$ & $0.129$
            & $\phantom{-}0.036$ $(0.018)$ & $0.045$ \\
        \texttt{lgbt}
            & $-0.095$ $(0.024)$ & $<0.001$
            & $\phantom{-}0.077$ $(0.024)$ & $0.002$ \\
        \texttt{womlab}
            & $-0.050$ $(0.018)$ & $0.005$
            & $\phantom{-}0.005$ $(0.019)$ & $0.783$ \\
        \texttt{plural}
            & $\phantom{-}0.119$ $(0.027)$ & $<0.001$
            & $\phantom{-}0.038$ $(0.028)$ & $0.176$ \\
        \texttt{anti-popul}
            & $-0.066$ $(0.018)$ & $<0.001$
            & $-0.025$ $(0.019)$ & $0.185$ \\
        \texttt{secular}
            & $-0.040$ $(0.022)$ & $0.073$
            & $-0.041$ $(0.022)$ & $0.062$ \\
        \texttt{rightwing}
            & $\phantom{-}0.221$ $(0.018)$ & $<0.001$
            & $-0.005$ $(0.017)$ & $0.751$ \\
        \midrule
        $R^2$             & \multicolumn{2}{c}{$0.220$} & \multicolumn{2}{c}{$0.059$} \\
        Adj.\ $R^2$       & \multicolumn{2}{c}{$0.219$} & \multicolumn{2}{c}{$0.057$} \\
        \bottomrule
    \end{tabular}
\end{table}
\clearpage

\clearpage

\end{document}